\documentclass[letterpaper, 10 pt, conference]{ieeeconf}  %

\IEEEoverridecommandlockouts                              %

\usepackage{booktabs}
\usepackage{xcolor}
\usepackage{pifont}
\usepackage[hidelinks]{hyperref}

\usepackage{amsmath} %
\usepackage{amssymb}  %
\usepackage{graphicx}
\usepackage{xspace}
\usepackage{glossaries}

\def\anonymous{0}   %

\if\anonymous0     %
    \newcommand{\ermanno}{Ermanno Bartoli}
    \newcommand{\ermannomail}{\texttt{bartoli@kth.se}}

    \newcommand{\buwei}{Buwei He}

    \newcommand{\dennis}{Dennis Rotondi}

    \newcommand{\seb}{Sebastian Koch}

    \newcommand{\federico}{Federico Tombari}

    \newcommand{\kai}{Kai O. Arras}

    \newcommand{\patric}{Patric Jensfelt}

    \newcommand{\yixi}{Yixi Cai}

    \newcommand{\iolanda}{Iolanda Leite}
    \newcommand{\iolandamail}{\texttt{iolanda@kth.se}}

    \newcommand{\rpl}{Department of Robotics, Perception and Learning}
    \newcommand{\kth}{KTH Royal Institute of Technology}
    \newcommand{\sweden}{Sweden}
    \newcommand{\stuttgart}{Socially Intelligent Robotics Lab at the Institute for Artificial Intelligence, University of Stuttgart, Germany}
    \newcommand{\mpi}{ Intl. Max Planck Research School for Intelligent Systems (IMPRS-IS)}
    \newcommand{\google}{Google Research}
\else
    \newcommand{\anonauthor}{Anonymous Author}
    \newcommand{\anonemail}{anonymous@anon.an}
    \newcommand{\anoninst}{Anonymous Institution}

    \newcommand{\buwei}{\anonauthor}

    \newcommand{\ermanno}{\anonauthor}
    \newcommand{\ermannomail}{\texttt{\anonemail}}
    
    \newcommand{\iolanda}{\anonauthor}
    \newcommand{\iolandamail}{\texttt{\anonemail}}

    \newcommand{\rpl}{\anoninst}
    \newcommand{\kth}{\anoninst}
    \newcommand{\sweden}{\anoninst}
\fi

\newacronym{rlhf}{RLHF}{Reinforcement Learning from Human Feedback}
\newacronym{cb}{CB}{Concept Bottleneck}
\newacronym{cbm}{CBM}{Concept Bottleneck Model}

\makeatletter
\renewcommand\paragraph{\@startsection{paragraph}{4}{\z@}%
  {0.8em plus 0.2em minus 0.2em}%
  {-1em}%
  {\normalfont\normalsize\bfseries}}
\makeatother
\newcommand{\modelname}{\textsc{GESTO}}
\newcommand{\xmark}{\ding{55}}

\title{\LARGE \bf

GESTO: Human-Centric Spatio-Temporal Memory for\\ Reasoning in Dynamic Scenes
} 

\author{%
\ermanno$^{*,1}$,
\buwei$^{*,1}$,
\dennis$^{2,3}$,
\seb$^{4}$,
\federico$^{4,5}$,\\
\kai$^{2}$,
\patric$^{1}$,
\yixi$^{1}$,
\iolanda$^{1}$%
\thanks{$^{*}$Equal contribution.}%
\thanks{$^{1}$\rpl, \kth, \sweden.
{\ermannomail}, {\iolandamail}}%
\thanks{$^{2}$\stuttgart}%
\thanks{$^{3}$\mpi}%
\thanks{$^{4}$\google}%
\thanks{$^{5}$TU Munich}%
}

\begin{document}

\maketitle
\thispagestyle{empty}
\pagestyle{empty}

\begin{abstract}
Robots operating in human environments need memories that capture not only
what objects exist and where, but also how people use them over time and how
individual interactions compose into goal-directed activities. Existing 4D
scene graphs preserve object and place histories but omit activity structure,
whereas activity representations are either not grounded in persistent 3D
scenes or rely on externally provided event boundaries and object
associations.
We present \modelname{} (Grounded Event and Spatio-Temporal memOry), a spatio-temporal memory that couples a persistent 4D
scene graph with a two-level hierarchy of atomic human--object interactions
and goal-driven events. From an RGB-D observation stream, \modelname{}
automatically extracts timestamped interactions, grounds them to persistent
scene entities, groups them into events, and uses event context to refine
uncertain object associations. A relation-aware tool-calling agent queries the
resulting memory for activity-centric spatio-temporal reasoning.
We evaluate \modelname{} on the reproducible text, binary, and time categories
of an existing benchmark, together with 40 new Space2Event and Event2Space
queries. \modelname{} achieves scores of 0.71, 0.75, and 0.70 on the
standard categories, approaching a method supplied with ground-truth event and
object grounding, while substantially outperforming the same reasoning
framework when these inputs are removed. It further achieves 0.73 and 0.75
on Space2Event and Event2Space queries. Ablations show that hierarchical event
structure and context-aware grounding refinement provide complementary
benefits, supporting activity-grounded hierarchical memory for retrospective
reasoning in dynamic human environments.
\end{abstract}

\section{Introduction}
Robots operating in human environments must reason not only about the current state of the world, but also about the activities that unfold within it. A service robot may be asked: ``Which mug was used to drink coffee this morning?'', ``Where does the person usually have breakfast?'', or ``Has this cup already been washed?'' Answering such questions requires more than recognizing objects and tracking their locations over time. They require a memory that records how people interacted with the environment, and how individual interactions fit into broader, goal-directed activities. 

Recent 4D scene graphs~\cite{gorlo2025describe, peddi2026spatiotemporalworldscenegraph, gorlo2026worth} have substantially improved the spatial and temporal memory of embodied agents. 
These representations maintain a queryable record of what objects were observed, where they were located, and how the scene evolved over time, but carry no notion of the activities that drove that evolution. 
Human behavior has been modeled with considerable structure, but along the following complementary lines which rarely meet. A substantial body of work in video understanding shows that activity is naturally hierarchical, decomposing goal-directed behavior into atomic actions and the higher-level activities they compose~\cite{gu2018ava,ji2020action,luo2021moma, song2023ego4dgoalstep}; this structure, however, is defined over frames or clips, with no link to the persistent objects and places a robot must act on. Representations that instead connect human behavior to a 3D scene take the opposite path: even the closest attempt, Event-Grounding Graph~\cite{nguyen2026event}, grounds events to spatial entities, but treats them as a flat set, with no composition into higher-level, goal-directed activities, and requires their temporal boundaries and object correspondences to be supplied externally rather than inferred from observation.

\begin{figure}
    \centering
    \includegraphics[width=0.8\linewidth]{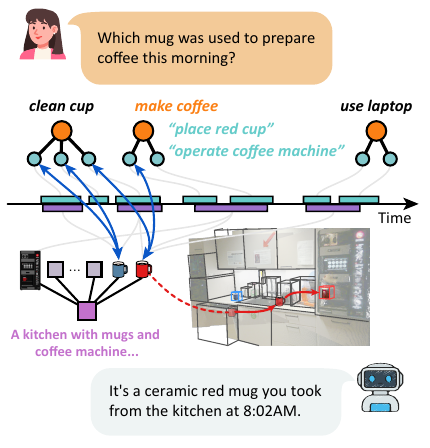}
    \caption{\modelname~answers retrospective queries about human behavior by reasoning over a grounded activity hierarchy. Given a natural language question, the agent traverses high-level events and atomic interactions linked to persistent objects and places in the 4D scene, returning an answer with supporting temporal and spatial evidence.}
    \label{fig:teaser}
    \vspace{-12pt}
\end{figure}

We present \modelname{} (\textbf{G}rounded \textbf{E}vent and \textbf{S}patio-\textbf{T}emporal mem\textbf{O}ry), a framework that augments 4D scene graphs with an activity hierarchy connecting two complementary levels of abstraction. 
The finer level holds \emph{atomic interactions}, short, timestamped human--object actions such as ``rinses cup'', and the coarser level groups related interactions into \emph{goal-driven events} such as ``prepares coffee''. Both levels are grounded to persistent objects and places in the scene graph through explicit links, so a query can move freely between geometry and activity semantics: from an object to the events in which it participated, from an event to the objects and locations it involved, and from a high-level event to the atomic interactions that support it. The entire representation is built automatically from an observation stream, without externally provided activity intervals, participating objects, or cross-event correspondences.

GESTO is designed as a memory representation for grounded retrospective reasoning. Accordingly, we evaluate it through the information that can be recovered from the constructed memory. Existing robotic 4D-scene benchmarks do not provide dense reference annotations for complete activity-grounded scene graphs, so we follow and extend the question-answering protocol introduced by EGG~\cite{nguyen2026event}.

Our contributions are threefold: (i) a human-centric spatio-temporal memory that augments a 4D scene graph with a two-level activity hierarchy of atomic human--object interactions and goal-driven events; (ii) a pipeline that constructs this representation autonomously from raw robot observations, including interaction extraction, event grouping, and grounding of activities to persistent scene entities; and (iii) an evaluation through a relation-aware tool-calling agent on activity-centric spatio-temporal question answering, matching methods that rely on ground-truth event and object grounding while constructing the memory fully automatically, across both robot and egocentric views. We open-source our prompts, queries and code upon acceptance.

\begin{figure*}[ht!]
    \centering
    \includegraphics[width=0.9\linewidth]{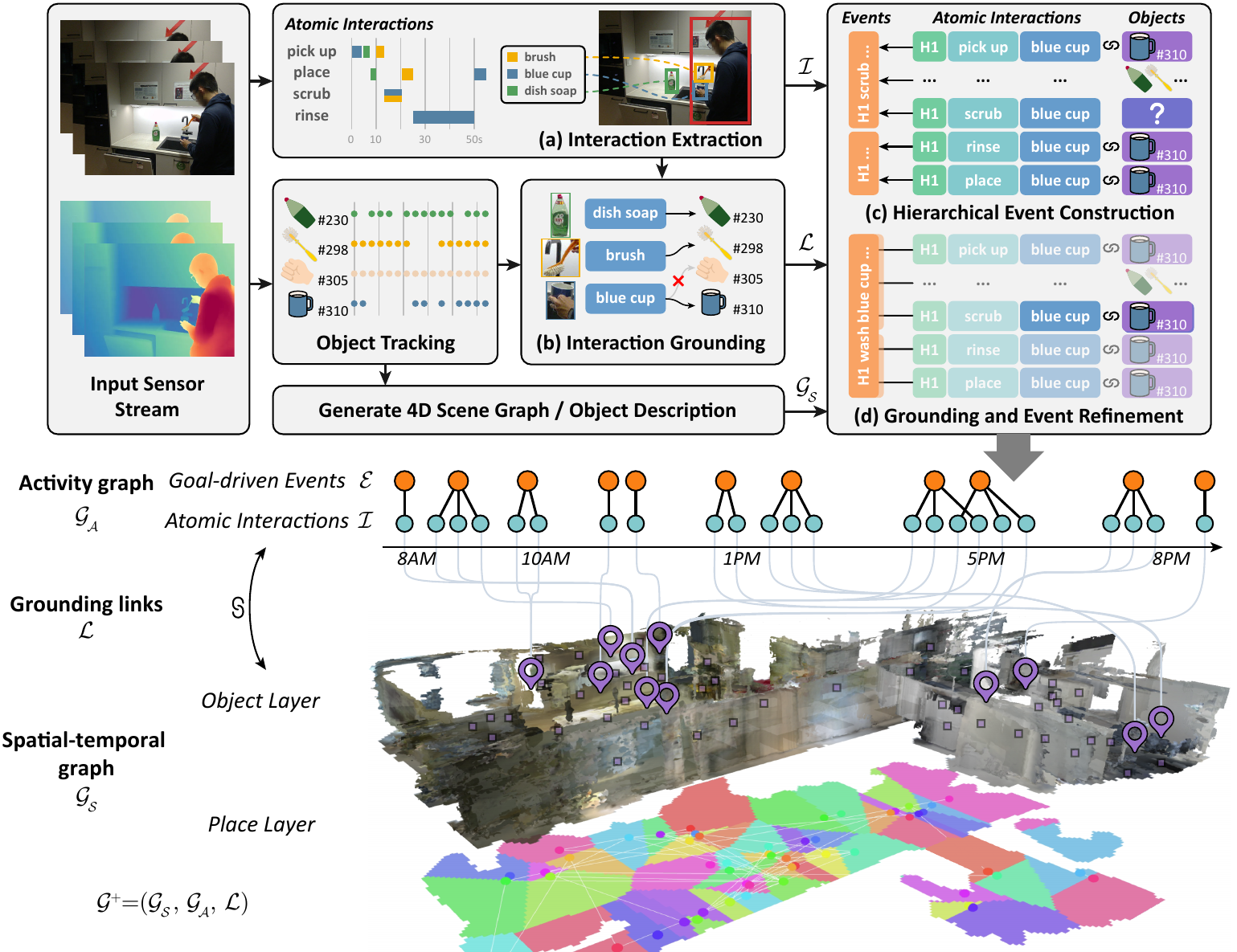}
    \caption{Overview of \modelname. (\textit{Left}) The online stage segments the RGB-D stream into short clips and extracts atomic human--object interactions, matching detections to tracked objects in $\mathcal{G}_{\mathrm{S}}$. Then, event grouping and context-aware grounding refinement organize atomic interactions into goal-driven events and resolve unmatched object links using activity context. (\textit{Right}) The resulting $\mathcal{G}^{+}$ jointly represents the spatio-temporal scene and the activity hierarchy, enabling queries that cross between geometry, objects, and human behavior over time.}
    \label{fig:method}
    \vspace{-12pt}
\end{figure*}

\section{Related work}

\subsection{Scene Graph Representations for Embodied Agents}
\textbf{3D scene graphs as spatial memory.}
3D scene graphs (3DSGs) have emerged as a compact, queryable representation that organizes metric and semantic information into a hierarchy of objects, places, and regions~\cite{armeni2019scenegraph,Werby-RSS-24,hughes2024foundations,werby2026keysg,rotondi2026survey}. Hydra~\cite{hughes2022hydra} showed that such graphs can be constructed incrementally and in real time; subsequent work enriches them with open-vocabulary semantics from foundation models~\cite{gu2024conceptgraphs,werby2024hierarchical,maggio2024clio} and with functional or affordance-level structure~\cite{zhang2025functional3dsg,rotondi2025fungraph,chang2025ashita}. These representations excel at describing \emph{what} objects are present and \emph{where}, but they are predominantly static and object-centric: they capture neither \emph{when} things happened nor the \emph{activities} that connect objects to the people who use them, limitations that dynamic and 4D scene graphs begin to address.

\textbf{Dynamic and 4D scene graphs.}
A growing body of work extends scene graphs along the temporal axis. Dynamic scene graphs introduce agent nodes that track moving humans through skeletal pose and trajectory~\cite{rosinol2020dynamic,rosinol2021kimera}; spatio-temporal SLAM systems such as Khronos~\cite{schmid2024khronos} jointly reason about short- and long-term scene change; and recent work abstracts temporal flow into 4D scene graphs~\cite{gorlo2025describe,catalano2025aion}. A complementary line stores per-frame or per-segment captions in retrieval databases~\cite{anwar2025remembr,xie2024embodied}, and episodic-memory approaches selectively retain only notable observations over long horizons~\cite{gorlo2026worth}. The closest to our setting, DAAAM~\cite{gorlo2025describe}, builds a hierarchical 4D scene graph with detailed open-vocabulary object descriptions and exposes it to a tool-calling agent for spatio-temporal question answering. Its representation is, however, organized around objects and observations: it records \emph{what} was seen, \emph{where} it was seen and \emph{when}, but not the \emph{activities} that connect objects to human behavior, which is precisely the gap our work addresses.

\subsection{Human-Centric and Hierarchical Activity Representations}
\textbf{Human-centric and activity-grounded representations.}
Modeling human behavior requires reasoning beyond individual object detections. Social 3D Scene Graphs~\cite{bartoli2025longtermplanning,bartoli2025social} augment static scenes with human--human and human--object interaction edges, encoding social and functional relations, but without temporal dynamics or any event structure above the level of a single interaction. Event-Grounding Graphs~\cite{nguyen2026event} take a step toward dynamic activity by linking events to spatial entities, demonstrating the value of spatially grounded events. Their event representation is, however, structurally flat: there is no hierarchy relating an individual interaction to a coarser episode, for instance, that a particular eating interaction is part of breakfast. Furthermore, their pipeline assumes externally provided activity intervals, participating objects, and object correspondences across events, an assumption we explicitly relax.

\textbf{Hierarchical event representations in video understanding.}
A substantial body of work from video understanding shows that human activity is inherently hierarchical. Cognitive studies confirm that people naturally segment continuous action into nested, multi-grain episodes~\cite{zacks2007event}, and early video models decompose goal-directed activities into recurring sub-units~\cite{kuehne2014language}. Modern datasets formalize this structure at increasing granularity, from atomic visual actions~\cite{gu2018ava} and frame-level interaction graphs~\cite{ji2020action} to multi-level activity hierarchies~\cite{luo2021moma,song2023ego4dgoalstep}, and graph-based video representations extend these ideas to dynamic and egocentric settings~\cite{yang2023psg4d,rodin2024easg,huang2025mindpalace}. The common limitation of these approaches is that their hierarchies are defined over frames or clips, with fixed taxonomies or offline annotations, and are not grounded in a persistent metric 3D environment. The structure exists in the video signal, but not in the world the robot navigates. A related body of work grounds activity recognition in egocentric video, localizing the objects and tools a person manipulates relative to the camera wearer \cite{grauman2022ego4d, damen2020epic, li2026egolive}. Because the camera is the actor's viewpoint, these representations are single-person by construction; a third-person observer such as a robot can instead relate the activities of several co-present people within one shared 3D memory.

In contrast, \modelname{} unifies these axes: it augments a dynamic 4D scene graph with an activity hierarchy that links timestamped atomic human--object interactions to higher-level events, grounds both levels to persistent scene entities, constructs this representation autonomously from raw robot observations, and queries it through a relation-aware tool-calling agent.

\section{Problem formulation}

Our objective is to construct a unified representation, $\mathcal{G}^{+}$, that jointly captures the dynamic spatial structure of the environment and the hierarchical temporal structure of human activities, by grounding activities in the objects and places they involve.

The representation is defined as
\begin{equation}
\mathcal{G}^{+} =
(\mathcal{G}_{\mathrm{S}},
\mathcal{G}_{\mathrm{A}},
\mathcal{L}),
\label{eq:repr}
\end{equation}
where $\mathcal{G}_{\mathrm{S}}$ is a spatio-temporal scene graph describing the environment, $\mathcal{G}_{\mathrm{A}}$ is a hierarchical activity graph describing human behaviour, and $\mathcal{L}$ is a set of grounding links connecting the two.

\noindent{\textbf{Spatio-temporal graph.}}
$\mathcal{G}_{\mathrm{S}}$ is a 4D scene graph that maintains persistent object nodes $\mathcal{O}$ and place nodes $\mathcal{P}$ over time, following the formulation introduced by Hydra~\cite{hughes2022hydra}. It represents \emph{what} entities exist, \emph{where} they are located, and \emph{when} they are observed, while each object additionally stores a semantic description of its appearance. Our framework treats $\mathcal{G}_{\mathrm{S}}$ as a modular backbone.
Consequently, any compatible 4D scene graph can be employed without modifying the remainder of the representation.

\noindent{\textbf{Activity graph.}}
$\mathcal{G}_{\mathrm{A}}$ is a hierarchical graph comprising two levels of abstraction: atomic interactions $\mathcal{I}$ and goal-driven events $\mathcal{E}$, representing activities.    

An atomic interaction $i=(h,o_{\textit{action}},a,\tau)\in\mathcal{I}$ represents a human $h$ performing an action $a$ on the involved interaction object $o_{\text{action}}$ during the time interval $\tau$, e.g., ``person rinses blue cup'' between 8:00 and 8:01 AM. 

A goal-driven event $\epsilon=(\mathcal{I}_{\epsilon}, s_{\epsilon})\in\mathcal{E}$
groups a coherent set of related interactions $\mathcal{I}_\epsilon\subseteq\mathcal{I}$ under the semantic summary $s_\epsilon$, e.g., ``prepares coffee''. The event interval $T_\epsilon$ spans from the start of the earliest constituent interaction to the end of the latest.

\noindent{\textbf{Grounding links.}}
The grounding links $\mathcal{L}$ couple the activity and spatial hierarchies at their finest level of granularity. Each link $\ell=(i,o)\in\mathcal{L}$ associates an atomic interaction $i\in\mathcal{I}$ with the persistent object node $o\in\mathcal{O}$. Since atomic interactions constitute goal-driven events and object nodes are organized under place nodes in $\mathcal{G}_{\mathrm{S}}$, these links connect the two hierarchies, allowing activities to be traced to the supporting spatial evidence and, conversely, spatial entities to be queried through the activities in which they participate.

The two graphs are complementary: $\mathcal{G}_{\mathrm{S}}$ captures objects but not the activities that connect them, $\mathcal{G}_{\mathrm{A}}$ captures activities but not where they occur, and $\mathcal{L}$ couples them along a shared session timeline. Of the three, $\mathcal{G}_{\mathrm{A}}$ and $\mathcal{L}$ are our contribution; $\mathcal{G}_{\mathrm{S}}$ is a swappable backbone.

\begin{figure}[h]
    \centering
    \includegraphics[width=0.85\linewidth]{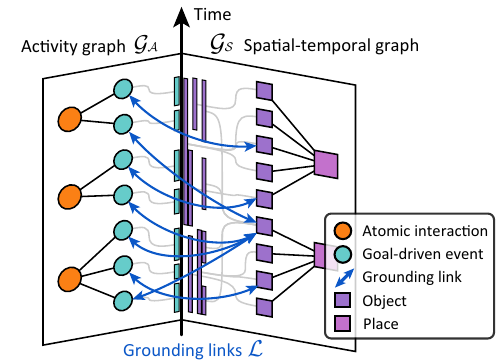}
    \caption{The coupled hierarchies of $\mathcal{G}^{+}$. The activity graph $\mathcal{G}_{\mathrm{A}}$ (left plane) organizes atomic interactions under goal-driven events, and the spatio-temporal graph $\mathcal{G}_{\mathrm{S}}$ (right plane) organizes persistent objects under places. The grounding links $\mathcal{L}$ connect them at their finest granularity, linking atomic interactions to object nodes, so that events inherit spatial context and objects become queryable through the activities they participate in.}
    \label{fig:planes}
    \vspace{-12pt}
\end{figure}

\section{Method}
\label{sec:method}

\modelname{} constructs $\mathcal{G}^{+}$ from an observation stream in two stages. First, it extracts atomic interactions, initializes their links to tracked objects using geometric and semantic evidence, and groups them into goal-driven events. It then uses event context to refine uncertain object links and re-evaluates the affected event assignments. The resulting memory is exposed to the relation-aware query agent described in Sec.~\ref{sec:querying}; Fig.~\ref{fig:method} summarizes the pipeline.

\subsection{Atomic Interaction}
\label{sec:interaction}

\noindent{\textbf{Interaction extraction.}}
While a person is in view, \modelname{} splits the streaming input into short, fixed-length video clips, and a vision-language model reads each to extract the atomic interactions it contains, following the granularity of atomic-action analysis in video~\cite{gu2018ava}.
For each interaction, the vision-language model produces (i) an interaction object label (e.g., ``blue cup'') and (ii) frame-wise bounding boxes over the interaction time window. Together, these observations provide the information required to associate the interaction with a persistent object node in the scene graph.

\noindent{\textbf{Interaction grounding.}}
After extraction, each interaction is then grounded to a persistent object node $o\in\mathcal{O}$ tracked by the scene-graph backbone (we use DAAAM~\cite{gorlo2025describe} in this work), to connect the activity and spatial representations.

The grounding combines geometric and semantic agreement. Geometric agreement is measured by the mean intersection-over-union (mIoU) between the localized boxes of $o_{\mathrm{action}}$ and the tracked boxes of $o$, while semantic agreement is measured by the text-embedding similarity between the interaction object label of $o_{\mathrm{action}}$ and the persistent object description of $o$. 
Among the candidate nodes whose mIoU exceeds $\tau_{\mathrm{mIoU}}$ and the semantic similarity exceeds $\tau_{\mathrm{sem}}$, the node with the highest semantic similarity is selected; candidates with borderline semantic similarity are further checked by an LLM judge.

The grounding cannot be finalized immediately because persistent object identities are still provisional and semantic descriptions are generated asynchronously. \modelname{} therefore first ranks candidate nodes online using mIoU, then finalizes the grounding once semantic descriptions are available and corrected when object nodes have been consolidated.
These established links form the measured grounding links $\mathcal{L}^{(0)}$, while interactions without reliable association remain temporarily unlinked until hierarchy contexts are available for further inference (Sec.~\ref{sec:refinement}).

\subsection{Goal-Driven Event}
\label{sec:event}

\noindent{\textbf{Hierarchical Event construction.}}
The upper level of $\mathcal{G}_{\mathrm{A}}$ groups interactions into the goal-driven events they compose: picking up a mug, operating the coffee machine, and filling it make up an event 
$\epsilon=(\mathcal{I}_{\epsilon},s_{\epsilon})\in\mathcal{E}$ 
with summary $s_{\epsilon}=$``prepares coffee''. An LLM performs the grouping under a few guiding principles: each event describes a single coherent goal at a consistent granularity, broader than an atomic interaction like ``picks up mug'' but not as sweeping as ``does kitchen work'', the level at which people naturally segment behavior~\cite{zacks2007event}; objects of the same category are distinguished by appearance, so similar objects are not conflated; and together the blocks $\{\mathcal{I}_{\epsilon}\}_{\epsilon\in\mathcal{E}}$
form a complete, time-ordered partition of the session's interactions
$\mathcal{I}$.
Composed this way, an event inherits the objects, places, and time span of its constituent interactions through the links established above, so it remains traceable to the timestamped evidence that supports it.

Applying this grouping to the interactions $\mathcal{I}$, together with their initial grounding links $\mathcal{L}^{(0)}$, produces an initial assignment of events $\mathcal{E}^{(0)}$.
The resulting event structure provides a higher-level activity context that can subsequently be used to revisit ambiguous or unresolved object links.

\subsection{Refinement}
\label{sec:refinement}

The grounding links $\mathcal{L}$ and event structure $\mathcal{E}$ are mutually informative.
Interactions belonging to the same goal-driven event provide context for resolving ambiguous object associations, while a corrected object association can in turn change whether an interaction remains consistent with its assigned event.
Starting from the initial estimates $\mathcal{L}^{(0)}$ and $\mathcal{E}^{(0)}$, \modelname{} performs one round of alternating refinement:
\begin{equation}
\begin{aligned}
\mathcal{L}^{(1)}
&=
\mathcal{R}_{\mathcal{L}}
\left(
\mathcal{L}^{(0)};\mathcal{E}^{(0)}
\right),
\\
\mathcal{E}^{(1)}
&=
\mathcal{R}_{\mathcal{E}}
\left(
\mathcal{E}^{(0)};\mathcal{L}^{(1)}
\right),
\end{aligned}
\label{eq:joint}
\end{equation}
where $\mathcal{R}_{\mathcal{L}}$ revises interaction-object links using the event structure as context, and $\mathcal{R}_{\mathcal{E}}$ subsequently re-evaluates event membership using the refined links.
The two refinement operators are described below.
We perform this alternation once rather than iterating Eq.~\eqref{eq:joint} to a fixed point.

\noindent{\textbf{Context-aware grounding refinement.}}
With the initial events $\mathcal{E}^{(0)}$ now available and providing structured contexts, in addition to the measured links $\mathcal{L}^{(0)}$, we can infer grounding links for interactions left unlinked due to insufficient association: we connect such an interaction to a nearby node only when the surrounding, already-grounded interactions and events agree on a single candidate, so no new node is invented and no ambiguous case is guessed. 
This creates the \emph{inferred} links $L_{\mathrm{inferred}}^{(1)}$ to update the links:
\begin{equation}
\mathcal L^{(1)} = \mathcal L^{(0)} \cup \mathcal L_{\mathrm{inferred}}^{(1)} .
\end{equation}

\noindent{\textbf{Event refinement.}}
For each interaction whose grounding was modified by
$\mathcal{R}_{\mathcal{L}}$, a lightweight LLM re-examines whether it remains coherent with its assigned event under the same principles used for initial event construction.
If the interaction is no longer consistent with that event, it will be reassigned to a compatible neighboring event; when no such event exists, a new event will be created for the interaction.

This step produces the refined event structure $\mathcal{E}^{(1)}$ while leaving $\mathcal{L}^{(1)}$ and $\mathcal{G}_{\mathrm{S}}$ unchanged, and constitutes the second update in Eq.~\eqref{eq:joint}.
We stop after this single alternating pass; iterating link and event refinement until mutual consistency is reached is left for future work.

\subsection{Relation-Aware Agent for Question-Answering}
\label{sec:querying}

We extend the object-centric tool-calling agent of DAAAM~\cite{gorlo2025describe} with tools that expose the activity-grounded memory $\mathcal{G}^{+}$, including events, constituent interactions, linked objects, places, and time intervals. These tools enables the agent to retrieve appropriate activity or spatial evidence according to the query. Answers are returned together with the supporting evidence.

\section{Experiments}

To evaluate \modelname, we conduct a series of experiments using our tool-calling agent and the proposed scene representation. We use the dataset introduced in~\cite{nguyen2026event}, as it captures an indoor environment in which a robot observes human activities over time from an RGB-D sensor stream. This setting is well-suited to evaluating our representation, which models the relationships between places, events, people, and objects over time.

\subsection{Experiment Setup}
\paragraph{Implementation details}
The observation stream is split into 15-second clips, from which Cosmos-Reason2~\cite{nvidia2026cosmosreason2_8b} extracts atomic interactions and localizes the manipulated object; the same model performs the event grouping and the borderline link judgments. Object masks are obtained with FastSAM~\cite{zhao2023fast}, and text embeddings are computed with sentence-t5-large~\cite{ni2022sentence}. We set $\tau_{\mathrm{mIoU}}=0.3$ and $\tau_{\mathrm{sem}}= 0.7$. Candidates with aggregated IoU below $\tau_{\mathrm{mIoU}}$ are discarded.

\paragraph{Standard benchmark}
We first evaluate \modelname~using the benchmark introduced in~\cite{nguyen2026event} to enable a direct comparison with prior work. The benchmark comprises 80 queries, grouped according to the expected answer format:
\textit{text} queries require a natural-language answer;
\textit{binary} queries require a true-or-false answer;
\textit{node} queries require the identification of one or more spatial nodes;
and \textit{time} queries require a timestamp.
For example, the benchmark includes questions such as ``Which room was the phone last seen used in?'', ``Was the red ceramic mug ever used to drink coffee?'', and ``What is the earliest time the person was seen working?''

We follow the evaluation protocol of~\cite{nguyen2026event}. Text answers are evaluated using an LLM-based semantic score between 0 and 1. Binary are evaluated using F1 scores. Time answers are considered correct if the predicted timestamp falls within two minutes of the ground-truth answer.

We found a small number of limitations in the benchmark from~\cite{nguyen2026event} worth noting, as they affect how some scores should be interpreted. Object descriptions are occasionally inconsistent between a question and its ground-truth answer: for instance, the same mug is referred to as the ``dark red mug'' in one question and the ``brown mug'' in its answer.
Since text and binary scoring compare predicted descriptions against the ground truth, such mismatches can penalize an otherwise correct answer. We also found that the benchmark includes an additional evaluation category, node queries. However, the ground-truth point clouds required for this evaluation are not publicly released, making it impossible to reproduce this category. We therefore exclude node queries from our comparison.
\paragraph{Extended queries}
Reasoning about human activity in 3D environments naturally calls for queries that cross between the spatial and activity domains: given a place or object, what happened there, and given an activity, where and with what did it occur. While the standard benchmark touches on these capabilities incidentally, it does not evaluate them as distinct categories. We therefore introduce two dedicated query sets, named \textbf{Space2Event} queries and \textbf{Event2Space} queries, of 20 queries each. Queries and ground-truth answers were written by manually inspecting the dataset recordings, before running any of the evaluated methods, and are scored with the same LLM-based semantic score used for text queries. \textbf{Space2Event} queries give a place or an object and ask for the activities associated with it, e.g. ``What is the red chair used for?'' or ``What happens in the lounge?''; \textbf{Event2Space} queries give an activity and ask for the places and objects involved, e.g., ``Which objects does the person use while working?'' or ``In which room do people usually mingle after lunch?''

\paragraph{Baselines and supervision settings}
We compare against DAAAM~\cite{gorlo2025describe}, ReMEmbR~\cite{anwar2025remembr}, and EGG~\cite{nguyen2026event}. DAAAM provides an object-centric 4D scene graph and ReMEmbR a long-horizon spatio-temporal memory, whereas EGG is the closest activity-centric baseline. The published EGG \textit{(auto-caption)} setting generates event descriptions automatically but retains externally provided activity intervals, participating objects, and cross-event object correspondences.

To evaluate EGG under the same fully automatic setting as \modelname{}, we also expose its tool-calling agent to the same automatically constructed 4D scene graph used to evaluate \modelname{}, including its predicted activity information. Thus, both methods operate on identical, imperfect scene and activity estimates, without externally provided event or object grounding; they differ in how this information is structured and exposed to the reasoning agent.

\begin{table*}[ht]
  \centering
  \caption{Comparison against EGG~\cite{nguyen2026event}, DAAAM~\cite{gorlo2025describe}, and ReMEmbR~\cite{anwar2025remembr} on the standard benchmark and our extended categories. EGG is reported with externally provided event and object grounding (\textit{auto-caption}) and fully automatic (\textit{w/o GT input}).}
  \label{tab:results}
  \begin{tabular}{lccccc}
  \toprule
  \textbf{Method} & \textbf{Text Acc.} & \textbf{Binary F1} & \textbf{Time Acc.}  & \textbf{Space2Event} & \textbf{Event2Space} \\
  \midrule
  DAAAM \cite{gorlo2025describe}              &  0.43      & 0.58       & 0.10    &  0.53      & 0.35        \\
  ReMEmbR \cite{anwar2025remembr}             & 0.52    & 0.49    & 0.38   & 0.46      & 0.41        \\
  EGG (auto-caption) \cite{nguyen2026event}  & 0.70    & \textbf{0.78}    & \textbf{0.78}  &   \xmark     &  \xmark \\
  EGG (w/o GT input) \cite{nguyen2026event}   & 0.33   & 0.29   & 0.50    &  0.60      & 0.50 \\
  \textbf{\modelname~(ours)}                                        & \textbf{0.71}     &  0.75  &  0.70  &  \textbf{0.73}      & \textbf{0.75}     \\
  \bottomrule
  \end{tabular}
  \end{table*}

\begin{table*}[ht]
  \centering
  \caption{Ablation of \modelname's components. We remove video fragmentation (forcing the VLM to process longer, unsegmented clips), context-aware grounding refinement (disabling the recovery of geometrically unmatched interactions), both, and the event hierarchy (retaining only atomic interactions, without grouping into goal-driven events).}
  \label{tab:ablation}
  \small
  \begin{tabular}{lccccc}
    \toprule
    \textbf{Method}
    & \textbf{Text Acc.}
    & \textbf{Binary F1}
    & \textbf{Time Acc.}
    & \textbf{Space2Event}
    & \textbf{Event2Space} \\
    \midrule
    \modelname~(w/o event hierarchy)
    & 0.52 & 0.71 & 0.40 & 0.53 & 0.65 \\

    \midrule
    \modelname~(w/o fragmentation)
    & 0.57 & 0.66 & 0.70 & 0.67 & 0.65 \\
    
    \modelname~(w/o refinement)
    & 0.55 & 0.69 & 0.59 & 0.65 & 0.61 \\
    
    \modelname~(w/o fragmentation \& refinement)
    & 0.53 & 0.66 & 0.50 & 0.63 & 0.55 \\
    
    \midrule
    \modelname~(full)
    & \textbf{0.71} & \textbf{0.75} & \textbf{0.70}
    & \textbf{0.73} & \textbf{0.75} \\
    \bottomrule
    \end{tabular}
    \vspace{-12pt}
\end{table*}

\subsection{Experiment Results}

Table~\ref{tab:results} shows that \modelname{} performs comparably to EGG
\textit{(auto-caption)}, despite constructing its activity representation
without externally provided activity intervals, participating objects, or
cross-event correspondences. \modelname{} achieves the highest text accuracy
($0.71$) and remains close to EGG \textit{(auto-caption)} on binary and time
queries ($0.75$ versus $0.78$, and $0.70$ versus $0.78$, respectively).
The largest gap is on time queries, which are particularly sensitive to
activity-recognition and object-grounding errors: confusing similar activities
or grounding an interaction to the wrong object can directly affect the
retrieved timestamp.

The two EGG settings highlight the impact of external grounding. With
automatically generated captions but externally provided activity intervals
and object associations, EGG performs strongly on the standard benchmark.
When its tool-calling agent is instead exposed to the same automatically
constructed, imperfect 4D scene and activity information used by \modelname{},
its performance drops to $0.33$ on text, $0.29$ on binary, and $0.50$ on time
queries. This fully automatic comparison therefore tests how the two reasoning
mechanisms use the same predicted evidence without externally supplied
grounding.

Compared with this setting, \modelname{} improves substantially across all
standard query types. Its activity hierarchy allows the agent to reason over
both grounded interactions and the goal-driven events they compose, retaining
information about what occurred, when, and with which persistent objects even
when individual observations are incomplete.

The strongest gains appear on Space2Event and Event2Space queries, which
explicitly require traversal between activity and spatial entities.
\modelname{} reaches $0.73$ and $0.75$, respectively, outperforming DAAAM, ReMEmbR and
fully automatic EGG. DAAAM maintains persistent objects and their spatial
history but lacks explicit human--object interaction structure, making it
difficult to distinguish objects that were merely nearby from those that were
actually used. These results support the benefit of grounding activity
structure to persistent objects and places.

EGG \textit{(auto-caption)} is not reported on the extended categories because
the released annotations lack the additional event--space associations required
by these queries; providing them manually would reintroduce the external
supervision that the fully automatic evaluation is intended to avoid.

On the standard benchmark, ReMEmbR outperforms DAAAM on text and time queries but is outperformed by \modelname{} on all three reported metrics.

\begin{figure}[ht!]
    \centering
    \includegraphics[width=0.8\linewidth]{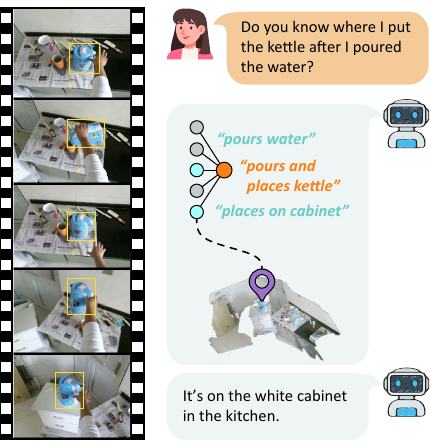}
    \caption{Qualitative demonstration of \modelname~on egocentric datasets. Given a natural-language query about a previously manipulated object, the agent retrieves the relevant interactions from $\mathcal{G}^{+}$ and returns a spatially grounded answer, using the same pipeline as the robot-mounted setting.}
    \label{fig:egocentric}
    \vspace{-12pt}
\end{figure}

\subsection{Ablation Studies}

\textbf{Video fragmentation.} Without splitting the stream into short clips, longer sequences are processed as a single input. Cosmos-Reason2 degrades noticeably once a clip exceeds its visual-token budget (approximately 40 seconds): it does not fail explicitly, but attends less precisely to individual interactions, producing coarser and less complete descriptions. Text accuracy ($0.71 \to 0.57$) and binary F1 ($0.75 \to 0.66$) drop accordingly, as both depend on identifying which interactions occurred and which objects were involved; time accuracy is unaffected, since even imprecise interactions retain enough temporal structure to locate \emph{when} an activity occurred.

\textbf{Context-aware grounding refinement.} 
Removing refinement leaves roughly 30\% of interactions without a geometric match, typically because of short temporal horizons or heavy self-occlusion. These interactions remain permanently ungrounded, which impacts temporal accuracy in particular ($0.70 \to 0.59$): when an ungrounded interaction is the only evidence for a temporal query, there is no grounded timestamp from which to answer. Space2Event and Event2Space show a similar drop, since they also rely on the grounding links that refinement recovers. Removing both fragmentation and refinement compounds these effects, resulting in the lowest text accuracy ($0.53$) and temporal accuracy ($0.50$).

\textbf{Event hierarchy.} Finally, we remove the event level entirely, leaving a flat set of grounded atomic interactions. This variant isolates the contribution of the hierarchy itself: perception and grounding are identical to the full system, yet time accuracy collapses to $0.40$ and Space2Event to $0.53$---on par with DAAAM, which has no activity representation at all. Without events, temporally scoped questions must be answered from dozens of unorganized interactions, and aggregating what a place or object is \emph{used for} loses the goal-level abstraction that summarizes it. Binary F1 ($0.71$) remains close to the full system, as many binary queries reduce to whether a single interaction occurred. The hierarchy, not merely the grounded interactions, is thus what carries the joint spatio-temporal reasoning.

Together, the ablations separate three failure modes: fragmentation preserves the resolution of interaction detection, refinement connects interactions to the persistent scene, and the event hierarchy organizes them into the structure that queries actually traverse.

\subsection{Egocentric Dataset}

While \modelname~is primarily evaluated on robot-mounted RGB-D sequences, the pipeline places no assumption on the camera configuration. We run the same pipeline on egocentric observations from the EgoLive and HOI4D dataset~\cite{li2026egolive,liu2022hoi4d}, adjusting only the Cosmos-Reason2 prompt to indicate a first-person perspective. Fig.~\ref{fig:egocentric} shows a representative example: given the query ``Do you know where I put the kettle after I poured the water?'', the agent traverses the grounded interactions \textit{pours water}, \textit{pours and places kettle}, \textit{places on cabinet}, and answers ``It's on the white cabinet in the kitchen''. While a quantitative egocentric evaluation is beyond our scope, the result illustrates that \modelname~transfers across sensor configurations without retraining, and systematic evaluation on egocentric benchmarks is an interesting future direction.

\section{Limitations}
Some limitations of the current pipeline point to directions for future work.

\textbf{Affordance-related capabilities.} The video-language model used for atomic interaction extraction performs well at recognizing what a person is doing, but degrades when asked to localize the specific object involved. This asymmetry arises because current VLMs rely on appearance-based features rather than functional or affordance-based reasoning, making them unreliable at grounding interactions to object instances under occlusion or when visually similar objects are present in the same frame. This affects the IoU-based matching step and is a limitation of the underlying model class rather than of our pipeline design.

\textbf{Semantic Mapping.} Reliable grounding also depends on the availability of accurate segmentation masks. When the person heavily occludes the manipulated object, masks may be missing entirely, leaving the interaction unlinked after the geometric grounding pass. When a mask is available but assigned to the wrong object, context-aware grounding refinement will attempt to correct it, but cannot always do so unambiguously. Both cases introduce noise into $\mathcal{G}^{+}$: unmatched interactions reduce coverage, mismatched links propagate incorrect associations to the event level. 

\textbf{Memory growth.} The memory grows with session length; detecting and consolidating recurrent activities (routines) is a natural direction for long-term deployments.

\section{Conclusions}
We presented \modelname, a representation that augments a 4D scene graph with a two-level activity hierarchy, grounding atomic human-object interactions and the goal-driven events they compose to persistent objects and places. The representation is built fully automatically from RGB-D observations, and queries through a relation-aware tool-calling agent.
Our results show that the benefit of activity-centric memory comes not merely
from storing additional descriptions, but from coupling persistent spatial
grounding with goal-level activity abstraction. Grounded interactions allow
the memory to distinguish objects that were actively used from those that were
only observed nearby, while goal-driven events organize local interactions
into units that support temporal and activity--space reasoning. Consequently,
\modelname{} remains competitive with methods given externally provided event
and object grounding and is substantially more robust when the representation
must be constructed fully automatically. These findings support hierarchical,
activity-grounded scene memory as a useful basis for retrospective reasoning
in dynamic human environments.

\bibliographystyle{IEEETran}
\bibliography{references}

\end{document}